\documentclass[10pt,twocolumn,letterpaper]{article}
\usepackage[compact]{titlesec}
\usepackage{microtype}
\usepackage{cvpr}
\usepackage{subfigure}
\usepackage{times}
\usepackage{epsfig}
\usepackage{graphicx}
\usepackage{amsmath}
\usepackage{amssymb}
\usepackage{bm}
\usepackage{multirow}
\usepackage{array, boldline, makecell, booktabs}
\usepackage{pifont}

 \newcommand{\compresslist}{%
 \setlength{\itemsep}{0pt}%
 \setlength{\parskip}{0pt}%
 \setlength{\parsep}{0pt}%
 }

\usepackage[pagebackref=true,breaklinks=true,letterpaper=true,colorlinks,bookmarks=false]{hyperref}

\cvprfinalcopy 

\def\cvprPaperID{17} 

\ifcvprfinal\fi

\begin{document}

\setlength{\textfloatsep}{8pt}

\setlength{\belowdisplayskip}{1pt} \setlength{\belowdisplayshortskip}{0pt}
\setlength{\abovedisplayskip}{1pt} \setlength{\abovedisplayshortskip}{0pt} 

\title{Holistically-Attracted Wireframe Parsing}
\author{{Nan Xue$^{1,2}$}, 
      {Tianfu Wu$^{2}$,
	 {Song Bai$^{3}$},
	 {Fudong Wang$^{1}$},
	 {Gui-Song Xia$^{1}$\thanks{Corresponding author}},
	 {Liangpei Zhang$^{1}$}, Philip H.S. Torr$^3$}
	 \vspace{1mm}\\
	 \and
	 {$^1$Wuhan University, China}
	 \\
	 {\small \tt \{xuenan, fudong-wang, guisong.xia, zlp62\}@whu.edu.cn}
    \and
    {$^2$NC State University, USA}
    \\
    {\small \tt tianfu\_wu@ncsu.edu}
	 \and
	 {$^3$University of Oxford, UK}\\
  	 {\small \tt songbai.site@gmail.com, philip.torr@eng.ox.ac.uk}
 }

\maketitle

\begin{abstract}
This paper presents a fast and parsimonious parsing method to accurately and robustly detect a vectorized wireframe in an input image with a single  forward pass. The proposed method is end-to-end trainable, consisting of three components: (i) line segment and junction proposal generation, (ii) line segment and junction matching, and (iii) line segment and junction verification. 
For computing line segment proposals, a novel exact dual representation is proposed which exploits a parsimonious  geometric reparameterization for line segments and forms a holistic 4-dimensional attraction field map for an input image. Junctions can be treated as the ``basins" in the attraction field. The proposed method is thus called Holistically-Attracted Wireframe Parser (HAWP). In experiments, the proposed method is tested on two benchmarks, the Wireframe dataset~\cite{Huang2018a} and the YorkUrban dataset~\cite{Denis2008}. On both benchmarks, it obtains state-of-the-art performance in terms of accuracy and efficiency. For example, on the Wireframe dataset, compared to the previous state-of-the-art method L-CNN~\cite{ZhouQM19}, it improves the challenging mean structural average precision (msAP) by a large margin ($2.8\%$ absolute improvements), and achieves 29.5 FPS on single GPU ($89\%$ relative improvement). 
A systematic ablation study is performed to further justify the proposed method.         
\end{abstract}

\vspace{-2mm}
\section{Introduction}
\vspace{-1mm}
\subsection{Motivations and Objectives}
\vspace{-2mm}
Line segments and junctions are prominent visual patterns in the low-level vision, and thus often used as important cues/features to facilitate many downstream vision tasks such as camera pose estimation~\cite{PribylZC15,PribylZC17,DuffKLP19}, image matching~\cite{asj-tip}, image rectification~\cite{XueXX019}, structure from motion (SfM)~\cite{BartoliS05,MicusikW17}, visual SLAM~\cite{Lemaire2007,ZhouZPYLY15,Zuo2017}, and surface reconstruction~\cite{LangloisBM19}. %
Both line segment detection and junction detection remain challenging problems in computer vision~\cite{XiaDG2014,afm,afm-pami}. Line segments and junctions are often statistically coupled in images. So, a new research task, \textbf{wireframe parsing}, is recently emerged to tackle the problem of jointly detecting meaningful and salient line segments and junctions with large-scale benchmarks available~\cite{Huang2018a}. And, end-to-end trainable approaches based on deep neural networks (DNNs) are one of the most interesting frameworks, which have shown remarkable performance.
\begin{figure}
    \centering
    \subfigure[Image\label{teaser:img}]{
    \includegraphics[width=0.31\linewidth]{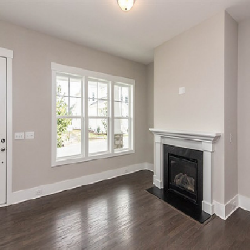}}
    \subfigure[Learned Lines\label{teaser:learned-ls}]{
    \includegraphics[width=0.31\linewidth]{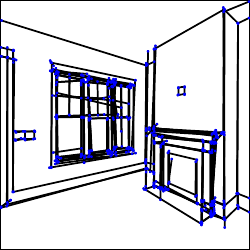}}
    \subfigure[HAWP (score$>$0.9)\label{teaser:HAWP}] {
    \includegraphics[width=0.31\linewidth]{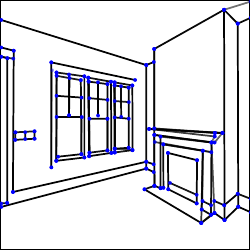}}
    \\
    \subfigure[Junction Proposals\label{teaser:junction}]{
    \includegraphics[width=0.31\linewidth]{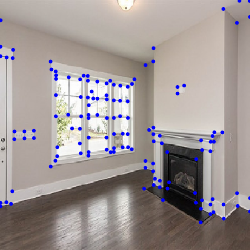}
    }
    \subfigure[Enumerated Lines\label{teaser:enumerated-ls}]{
    \includegraphics[width=0.31\linewidth]{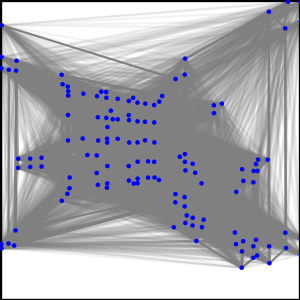}}
    \subfigure[L-CNN (score$>$0.9)\label{teaser:L-CNN}] {
    \includegraphics[width=0.31\linewidth]{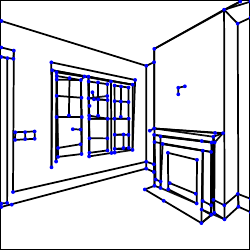}}
    \caption{\small Illustration of the proposed HAWP in comparison with L-CNN~\cite{ZhouQM19} in wireframe parsing. The two methods adopt the same two-stage parsing pipeline: proposal (line segments and junctions) generation and proposal verification. They use the same junction prediction in \subref{teaser:junction} and verification modules. The key difference lies in the line segment proposal generation. L-CNN bypasses directly learning line segment prediction module and resorts to a sophisticated sampling based approach for generation line segment proposals in \subref{teaser:enumerated-ls}. Our HAWP proposes a novel line segment prediction method in \subref{teaser:learned-ls} for more accurate and efficient parsing,~\eg,~the parsing results of the window in \subref{teaser:HAWP} and \subref{teaser:L-CNN}.}
    \vspace{3mm}
    \label{fig:teaser}
\end{figure}

In wireframe parsing, it can be addressed relatively better to learn a junction detector with state-of-the-art deep learning approaches and the heatmap representation (inspired by its widespread use in human pose estimation~\cite{NewellYD16,VarolCRYYLS18,ZhouZPLDD19}). This motivated a conceptually simple yet powerful wireframe parsing algorithm called L-CNN~\cite{ZhouQM19}, which achieved state-of-the-art performance on the Wireframe benchmark~\cite{Huang2018a}. L-CNN bypasses learning a line segment detector. It develops a sophisticated and carefully-crafted sampling schema to generate line segment proposals from all possible candidates based on the predicted junctions, and then utilizes a line segment verification module to classify the proposals. A large number of proposals are entailed for achieving good results at the expense of computational costs. And, ignoring line segment information in the proposal stage may not take full advantage of the deep learning pipeline for further improving performance.

On the other hand, without leveraging junction information in learning, the recently proposed attraction field map (AFM) based approaches~\cite{afm,afm-pami} are the state-of-the-art methods for line segment detection. 
AFM is not strictly end-to-end trainable. The reparameterization of pixels in the lifting process is for lines, instead of line segments (\ie,~we can only infer a line with a given displacement vector, and that is why the squeezing module is needed). 

In this paper, we are interested in learning an end-to-end trainable and fast wireframe parser. First, we aim to develop an exact dual and parsimonious reparameterization scheme for line segments, in a similar spirit to the AFM~\cite{afm}, but without resorting to the heuristic squeezing process in inference. Then, we aim to tackle wireframe parsing by leveraging both line segment and junction proposals to improve both accuracy and efficiency and to eliminate the carefully-crafted sampling schema as done in L-CNN~\cite{ZhouQM19}. 

\subsection{Method Overview}
\begin{figure}
    \centering
    \includegraphics[width=0.95\linewidth]{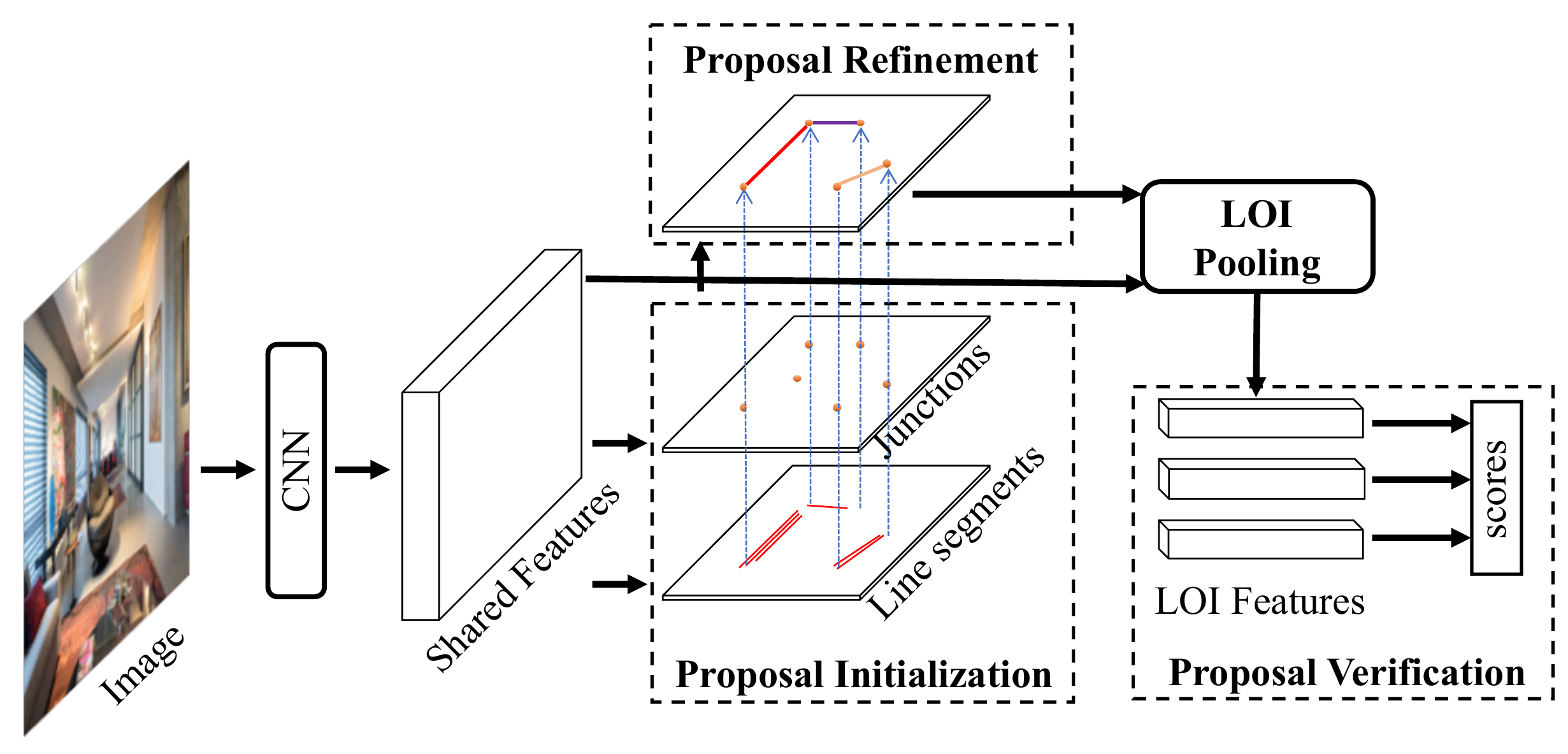}
    \caption{Illustration of the architecture of our proposed HAWP. It consists of three components, proposal initialization, proposal refinement and proposal verification. See text for details.}
    \label{fig:network}
\end{figure}

\vspace{-2mm}
In general, a parsing algorithm adopts two phases as proposed in the generic image parsing framework~\cite{TuParsing}: proposal generation and proposal verification, which are also realized in the state-of-the-art object detection and instance segmentation framework~\cite{fastRCNN,fasterrcnn,maskrcnn}. 
The current state-of-the-art wireframe parser, L-CNN~\cite{ZhouQM19} follows the two-phase parsing paradigm. The proposed method in this paper also adopts the same setup. As illustrated in Fig.~\ref{fig:teaser} and Fig.~\ref{fig:network}, the proposed method consists of three components: 

\textit{i) Proposal initialization: line segment detection and junction detection}. Given an input image, it first passes through a shared feature backbone (\eg, the stacked Hourglass network~\cite{NewellYD16}) to extract deep features. Then, for junction detection, we adopt the same head regressor based on the heatmap representation as done in L-CNN~\cite{ZhouQM19} (Section~\ref{sec:juncdet}), from which the top-$K$ junctions are selected as initial junction proposals. For computing line segment proposals, a novel method is proposed (Section~\ref{sec:lsd}).

\textit{ii) Proposal refinement: line segment  and junction matching}. The matching is to calculate meaningful alignment between line segment initial proposals and junction initial proposals. In the refinement (Section~\ref{sec:matching}), a line segment proposal is kept if its two end-points are supported by two junction proposals. If a junction proposal does not find any support line segment proposal, it will be removed. 

\textit{iii) Proposal verification: line segment and junction classification}. The verification process is to classify (double-check) the line segments and junctions from the proposal refinement stage. We utilize the same verification head classifier (Section~\ref{sec:verification}) as done in  L-CNN~\cite{ZhouQM19}, which exploits a Line-of-Interest (LOI) pooling operation to compute features for a line segment, motivated by the Region-of-Interest (ROI) pooling operation used in the popular two-stage R-CNN frameworks~\cite{fastRCNN,fasterrcnn,maskrcnn}.        

Geometrically speaking, the proposed wireframe parser is enabled by the holistic 4-D attraction field map and the ``basins" of the attraction field revealed by junctions. We thus call the proposed method a \textbf{Holistically-Attracted Wireframe Parser (HAWP)}.
The proposed HAWP is end-to-end trainable and computes a vectorized wireframe for an input image in single forward pass. The key difference between our HAWP and the current state-of-the-art L-CNN~\cite{ZhouQM19} approach is the novel line segment reparameterization and its end-to-end integration in the parsing pipeline. Our HAWP outperforms L-CNN by a large margin in terms of both accuracy and efficiency (Section~\ref{sec:exp}).

\section{Related Work and Our Contributions}
\vspace{-2mm}
The fundamental problem in wireframe parsing is to learn to understand the basic physical and geometric constraints of our world. The problem can date back to the pioneering work of understanding Blocks World by Larry Roberts~\cite{roberts1963machine,GuptaEH10} at the very beginning of computer vision. We briefly review two core aspects as follows.

\textbf{Representation of Line Segments.} There is a big gap between the mathematically simple geometric representation of line segments (at the symbol level) and the raw image data (at the signal level). A vast amount of efforts have been devoted to closing the gap with remarkable progress achieved. Roughly speaking, there are three-level representations of line segments developed in the literature: (i)\textit{ Edge-pixel based representations}, which are the classic approaches and have facilitated a tremendous number of line segment detectors~\cite{Ballard81,ppht,BurnsHR86,DesolneuxMM00,VonGioi2010,Cho2018,VonGioi2010,Almazan_2017_CVPR,Cho2018}. Many of these line segment detectors suffer from two fundamental issues inherited from the underlying representations: the intrinsic uncertainty and the fundamental limit of edge detection, and the lack of structural information guidance from pixels to line segments. The first issue has been eliminated to some extent by state-of-the-art deep edge detection methods~\cite{HED-IJCV,COB-PAMI}. (ii) \textit{Local support region based representations}, \eg, the level-line based support region used in the popular LSD method~\cite{VonGioi2010} and its many variants~\cite{AkinlarT11,Cho2018}. The local support region is still defined on top of local edge information (gradient magnitude and orientation), thus inheriting the fundamental limit. (iii) \textit{Global region partition based representation}, which is recently proposed in the AFM method~\cite{afm}. AFM does not depend on edge information, but entails powerful and computationally efficient DNNs in learning and inference. AFM is not strictly an exact line segment representation, but a global region partition based line representation. \textit{The issue is addressed in this paper by proposing a novel holistic AFM representation} that is parsimonious  and exact for line segments.        

\textbf{Wireframe Parsing Algorithm Design.} The recent resurgence of wireframe parsing, especially in an end-to-end way, is driven by the remarkable progress of DNNs which enables holistic map-to-map prediction (\eg, from raw images to heatmaps directly encoding edges~\cite{HED-IJCV} or human keypoints~\cite{ConvPose}, \etc). As aforementioned, the general framework of parsing is similar between different parsers. Depending on whether line segment representations are explicitly exploited or not, the recent work on wireframe parsing can be divided into two categories: (i) \textit{Holistic wireframe parsing}, which include data-driven proposal generation for both line segments and junctions, \eg, the deep wireframe parser (DWP)~\cite{Huang2018a} presented along with the wireframe benchmark. DWP is not end-to-end trainable and relatively slow. (ii) \textit{Deductive wireframe parsing}, which  utilizes data-driven proposals only for junctions and resorts to sophisticated top-down sampling methods to deduce line segments based on detected junctions,~\eg,~PPG-Net~\cite{ZhangLBZWHKXG19} and L-CNN~\cite{ZhouQM19}. The main drawbacks of deductive wireframe parsing are in two-fold: high computational expense for line segment verification, and over-dependence on junction prediction. \textit{The proposed HAWP is in the first category, but enjoys end-to-end training and real-time speed.}  

\textbf{Our Contributions.}
This paper makes the following main contributions to the field of wireframe parsing:
\begin{enumerate}\compresslist
    \item[-] It presents a novel holistic attraction field to exactly characterize the geometry of line segments. To our knowledge, this is the first work that facilitates an exact dual representation for a line segment from any distant point in the image domain and that is end-to-end trainable.
    
    \item[-] It presents a holistically-attracted wireframe parser (HAWP) that extracts vectorized wireframes in input images in a single forward pass. 
    
    \item[-] The proposed HAWP achieves state-of-the-art performance (accuracy and efficiency) on the Wireframe dataset~\cite{Huang2018a} and the YorkUrban dataset~\cite{Denis2008}. 
\end{enumerate}
\section{Holistic Attraction Field Representation}\label{sec:HAFM}
\vspace{-2mm}
In this section, we present the details of our proposed holistic attraction field  representation of line segments. The goal is to develop an exact dual representation using geometric reparameterization of line segments, and the dual representation accounts for non-local information and enables leveraging state-of-the-art DNNs in learning. By an exact dual representation, it means that in the ideal case it can recover the line segments in closed form. Our proposed holistic attraction field representation is motivated by, and generalizes the recent work called attraction field map (AFM)~\cite{afm}. 

We adopt the vectorized representation of wireframes in images~\cite{Huang2018a}, that is we use real coordinates for line segments and junctions, rather than discrete ones in the image lattice. Denote by $\Lambda$ and $\mathcal{D}\subset \mathbb{R}^2$ the image lattice (discrete) and the image domain (continuous) respectively.  A line segment is denoted by its two end-points, $\ddot{l}=(\mathbf{x}_1, \mathbf{x}_2)$, where $\mathbf{x}_1, \mathbf{x}_2\in \mathcal{D}$ (2-D column vector). The corresponding line equation associated with $\ddot{l}$ is defined by,  $l:  \mathbf{a}^T_{\ddot{l}} \cdot \mathbf{x} + b_{\ddot{l}} = 0$ where $\mathbf{a}_{\ddot{l}}\in \mathbb{R}^2$ and $b_{\ddot{l}}\in \mathbb{R}$, and they can be solved in closed form given the two end-points.  

\textbf{Background on the AFM method}~\cite{afm}. To be self-contained, we briefly overview the AFM method. The basic idea is to ``\textit{lift}" a line segment to a region, which facilitates leveraging state-of-the-art DNNs in learning. To compute the AFM for a line segment map, each (pixel) point $\mathbf{p}\in \Lambda$ is assigned to a line segment $\ddot{l}$ if it has the minimum distance to $\ddot{l}$ among all line segments in a given image. The distance is calculated as follows. Let $\mathbf{p}'$ be the point projected onto the line $l$ of a line segment $\ddot{l}$. If $\mathbf{p}'$ is not on the line segment $\ddot{l}$ itself, it will be re-assigned to one of the two end-points that has the smaller Euclidean distance. Then, the distance between $\mathbf{p}$ and $\ddot{l}$ is the Euclidean distance between $\mathbf{p}$ and $\mathbf{p}'$. If $\mathbf{p}$ is assigned to $\ddot{l}$, it is reparameterized as $\mathbf{p}-\mathbf{p}'$, \ie, the displacement vector in the image domain. The AFM of a line segment map is a 2-D vector field, which is created by reparameterizing all the (pixel) points in the image lattice $\Lambda$ and often forms a region partition of the image lattice. A heuristic squeezing module is also proposed in the AFM work to recover a line segment from a 2-D vector field region (\emph{a.k.a.}, attraction). 

\begin{figure}
    \centering
    \includegraphics[width=0.99\linewidth]{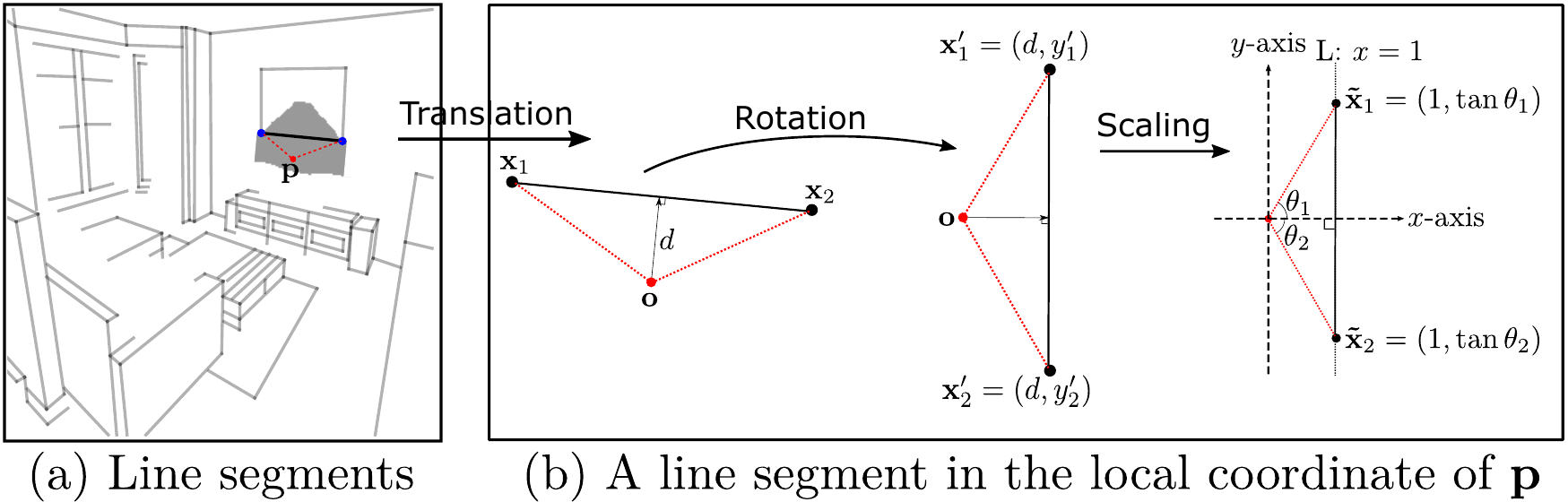}
    \caption{An illustration for representing line segments in images with the related distant points. (a) shows one of the line segments (marked black with two blue endpoints), the corresponding support region (marked gray) calculated by AFM~\cite{afm} and one of the distant points in the support region. (b) shows the process of extending the attraction field representation and transforming the line segment into a standard local coordinate originated at $\mathbf{p}$ with a horizontal unit attraction vector. }
    \label{fig:HAF}
\end{figure}

\textbf{The proposed holistic attraction field map.} Strictly speaking, the displacement vector based geometric reparameterization scheme in the AFM method can only provide complete information for the underlying line $l$ of a line segment $\ddot{l}$ (when the projection is not outside the line segment).
One \textit{straightforward extension } of the AFM method is as follows. As illustrated in the first column in Fig.~\ref{fig:HAF} (b), consider a distant (pixel) point $\mathbf{p}$ outside a line segment $\ddot{l}$ with the projection point being on the line segment, if we not only use the displacement vector between $\mathbf{p}$ and its projection point, but also include the two displacement vectors between $\mathbf{p}$ and the two end-points of the line segment, we can reparameterize $\mathbf{p}$ by its 6-D displacement vector which can completely determine the line segment (\ie, an exact dual representation). There are some points (pixels) (\eg, points on any line segment) that should not be reparameterized to avoid degradation and are treated as the ``background". Thus, we can create a 6-D attraction field and each line segment is supported by a region in the field map (shown by the gray region in Fig.~\ref{fig:HAF} (a)).  This was our first attempt in our study, and it turns out surprisingly that the 6-D attraction field can not be accurately and reliably learned in training with deep convolutional neural networks. We hypothesis that although the 6-D attraction field captures the sufficient and necessary information for recovering line segments in closed form, it is not parsimoniously and complementarily encoded using 3 displacement vectors for each point, which may increase the  difficulty of learning even with powerful DNNs.         

\textit{We derive an equivalent geometric encoding that is parsimonious  and complementary } as shown in the right two columns in Fig.~\ref{fig:HAF}. For a line segment $\ddot{l}$, our derivation undergoes a simple affine transformation for each distant pixel point $\mathbf{p}$ in its support region. Let $d$ be the distance between $\mathbf{p}$ and $\ddot{l}$, \ie, $d=|\mathbf{a}^T_{\ddot{l}}\cdot \mathbf{p}' + b_{\ddot{l}}|>0$. We have,  
\begin{enumerate}\compresslist
    \item [i)] \textit{Translation}: The point $\mathbf{p}$ is then used as the new coordinate origin. 
    \item [ii)] \textit{Rotation}: The line segment is then aligned with the vertical $y$-axis with the end-point $\mathbf{x}_1$ on the top and the point $\mathbf{p}$ (the new origin) to the left. The rotation angle is denoted by $\theta\in [-\pi, \pi)$.
    \item [iii)] \textit{Scaling}: The distance $d$ is used as the unit length to normalize the $x$- / $y$-axis in the new coordinate system.  
\end{enumerate}

In the new coordinate system after the affine transformation, let $\theta_1$ and $\theta_2$ be the two angles as illustrated in Fig.~\ref{fig:HAF} ($\theta_1\in (0, \frac{\pi}{2})$ and $\theta_2\in (-\frac{\pi}{2}, 0]$). So, a point $\mathbf{p}$ in the support region of a line segment $\ddot{l}$ is reparameterized as, 
\begin{equation}
    \mathbf{p}(\ddot{l}) = (d, \theta, \theta_1, \theta_2), \label{eq:HAF}
\end{equation}
which is completely equivalent to the 6-D displacement vector based representation and thus capable of recovering the line segment in closed form in the ideal case. 
For the ``background" points which are not attracted by any line segment based on our specification, we encode them by a dummy 4-D vector $(-1, 0, 0, 0)$. 

The derived 4-D vector field map for a line segment map is called a \textbf{holistic attraction field map} highlighting its completeness and parsimoniousness for line segments, compared to the vanilla AFM~\cite{afm}. 

\textbf{High-level explanations of why the proposed 4-D holistic AFM is better than the 6-D vanilla AFM.} Intuitively, for a line segment and a distant point $\mathbf{p}$, we can view the support region (the grey one in Fig.~\ref{fig:HAF} (a)) as ``a face" with the point $\mathbf{p}$ being the left ``eye" center and the line segment being the vertical ``head bone". So, the affine transformation stated above is to ``align" all the ``faces"~\emph{w.r.t.}~the left ``eye" in a canonical frontal viewpoint. It is well-known that this type of  ``representation normalization" can eliminate many nuisance factors in data to facilitate more effective learning. Furthermore, the joint encoding that exploits displacement distance and angle effectively decouples the attraction field~\emph{w.r.t.}~complementary spanning dimensions.      

\section{Holistically-Attracted Wireframe Parser}\label{sec:HAWP}
\vspace{-2mm}
In this section, we present details of our Holistically-Attracted Wireframe Parser (HAWP). 

\textbf{Data Preparation.}~Let $D_{train}=\{(I_i, L_i); i=1, \cdots, N\}$ be the set of training data where all the images $I_i$'s are resized to the same size of $\Lambda=H\times W$ pixels, and $L_i$ is the set of $n_i$ annotated line segments in the image $I_i$, $L_i=\{\ddot{l}_{i,1}, 
\cdots, \ddot{l}_{i,n_i}\}$ and each line segment $\ddot{l}_{i, j}=(\mathbf{x}_{i,j,1}, \mathbf{x}_{i,j,2})$ is represented by its two annotated end-points (the vectorized wireframe representation). 

\textit{The groundtruth junction heatmap representations.} We adopt the same settings used in L-CNN~\cite{ZhouQM19}. 
For an image $I\in D_{train}$ (the index  subscript is omitted for simplicity), the set of unique end-points from all line segments are the junctions, denoted by $J$. Then, we create two maps: the junction mask map, denoted by $\mathcal{J}$, and the junction 2-D offset map, denoted by $\mathcal{O}$. A coarser resolution is used in computing the two maps by dividing the image lattice into $H'\times W'$ bins (assuming all bins have the same size, $B\times B$,~\ie,~the down-sampling rate is $B=\frac{H}{H'}=\frac{W}{W'}$). Then, for each bin $b$, let $\Lambda_b\subset \Lambda$ and $\mathbf{x}_b\in \Lambda_b$ be its corresponding patch and the center of the patch respectively in the original image lattice and we have,   
$
    \mathcal{J}(b) = 1 \text{ and } \mathcal{O}(b) = 
    (\mathbf{x}_b - \mathbf{p}) \text{ if }\exists \mathbf{p}\in J, \text{ and } \mathbf{p}\in \Lambda_{b}
$
and both are set to 0 otherwise, where the offset vector in $\mathcal{O}(b)$ is normalized by the bin size, so the range of $\mathcal{O}(b)$ is bounded by $[-\frac{1}{2}, \frac{1}{2})\times [-\frac{1}{2}, \frac{1}{2})$.

\textit{The groundtruth holistic attraction field map.}~It is straightforward to follow the definitions in Section~\ref{sec:HAFM} to compute the map for an image $I\in D_{train}$. Denote by $\mathcal{A}$ be the map of the size $H'\times W'$ (the same as that of the two junction maps), which is initialized using the method in Section~\ref{sec:HAFM}. Then, we normalize each entry of the 4-D attraction field vector (Eqn.~\eqref{eq:HAF}) to be in the range $[0, 1)$. We select a distance threshold $d_{max}$. We filter out the points in $\mathcal{A}$ if their $d$'s are greater than $d_{max}$ by changing them to the ``background" with the dummy vector $(-1, 0, 0, 0)$. Then, we divide the distances (the first entry) of the remaining non-background points by $d_{max}$. Here, $d_{max}$ is chosen such that all line segments still have sufficient support distant points ($d_{max}= 5$ in our experiments). It also helps remove points that are far away from all line segments and thus may not provide meaningful information for LSD. For the remaining three entries, it is straightforward to normalize based on their bounded ranges. For example, an affine transformation is used to normalize $\theta$ to $\frac{\theta}{2\pi}+\frac{1}{2}$.   

\textbf{Feature Backbone.} We chose the stacked Hourglass network~\cite{NewellYD16} which is widely used in human keypoint estimation and corner-point based object detection~\cite{CornerNet}, and also adopted by L-CNN~\cite{ZhouQM19}. The size of the output feature map is also $H'\times W'$. Denote by $\mathcal{F}$ the output feature map for an input image $I$.    

\subsection{Computing Line Segment Proposals}\label{sec:lsd}
\vspace{-2mm}
Line segment proposals are computed by predicting the 4-D AFM $\mathcal{A}$ from $\mathcal{F}$. Let $\hat{\mathcal{A}}$ be the predicted 4-D map.  $\hat{\mathcal{A}}$ is computed by 
an 
$1\times 1$ convolutional layers followed by a sigmoid layer. With $\hat{\mathcal{A}}$, it is straightforward to generate line segment proposals by reversing the simple normalization step and the geometric affine transformation (Section~\ref{sec:HAFM}). However, we observe that the distance (the first entry) is more difficult to predict in a sufficiently accurate way. We leverage an auxiliary supervised signal in learning, which exploits the distance residual, in a similar spirit to the method proposed for depth prediction in~\cite{ChenHXS19}. In addition to predict $\hat{\mathcal{A}}$ from $\mathcal{F}$, we also compute a distance residual map, denoted by $\hat{\Delta \mathbf{d}}$, using one $1\times 1$ convolutional layers followed by a sigmoid layer. The groundtruth for $\hat{\Delta \mathbf{d}}$, denoted by $\Delta \mathbf{d}$, is computed by the residual (the absolute difference) between the two distances in $\mathcal{A}$ and $\hat{\mathcal{A}}$ respectively.

In training, 
channel-wise $\ell_1$ norm
is used as the loss function for both $\mathbb{L}(\mathcal{A}, \hat{\mathcal{A}})$ and $\mathbb{L}(\Delta \mathbf{d}, \hat{\Delta \mathbf{d}})$. The total loss for computing line segments is the sum of the two losses, $\mathbb{L}_{LS}=\mathbb{L}(\mathcal{A}, \hat{\mathcal{A}})+\mathbb{L}(\Delta \mathbf{d}, \hat{\Delta \mathbf{d}})$. In inference, with the predicted $\hat{d}\in \hat{\mathcal{A}}$ and $\hat{\Delta d}\in \hat{\Delta \mathbf{d}}$ (both are non-negative due to the sigmoid transformation), since we do not know the underlying sign of the distance residual, we enumerate three possibilities in updating the distance prediction, 
\begin{equation}
    \hat{d}'(\kappa) = \hat{d} + \kappa \cdot \hat{\Delta d}, 
\end{equation}
where $\kappa=-1, 0, 1$. So, each distant point may generate up to three line segment proposals depending on whether the condition  $0< \hat{d}'(\kappa) \leq d_{max}$ is satisfied. 

\subsection{Junction Detection}\label{sec:juncdet}
\vspace{-2mm}
Junction detection is addressed by predicting the two maps, the junction mask map and the junction offset map, from the feature map $\mathcal{F}$. They are computed by  one $1\times 1$ convolutional layers followed by a sigmoid layer. Denote by $\hat{\mathcal{J}}$ and $\hat{\mathcal{O}}$ the predicted mask map and offset map respectively. The sigmoid function for computing the offset map has an intercept $-0.5$.
In training, the binary cross-entropy loss is used for $\mathbb{L}(\mathcal{J}, \hat{\mathcal{J}})$, and the $\ell_1$ loss  is used for $\mathbb{L}(\mathcal{O}, \hat{\mathcal{O}})$, following the typical setting in heatmap based regression for keypoint estimation tasks and consistent with the use in L-CNN~\cite{ZhouQM19}. The total loss is the weighted sum of the two losses, $\mathbb{L}_{Junc}=\lambda_{msk}\cdot \mathbb{L}(\mathcal{J}, \hat{\mathcal{J}}) + \lambda_{off}\cdot \mathcal{J} \odot \mathbb{L}(\mathcal{O}, \hat{\mathcal{O}})$, where $\odot$ represents element-wise product, and $\lambda_{msk}$ and $\lambda_{off}$ are two trade-off parameters (we set $\lambda_{msk}$ and $\lambda_{off}$ to $8.0$ and $0.25$ respectively in our experiments). In inference, we also apply the standard non-max suppression (NMS)~\emph{w.r.t.}~a $3\times 3$ neighborhood, which can be efficiently implemented by a modified max-pooling layer. After NMS, we keep the top-$K$ junctions from $\hat{\mathcal{J}}$. 
And, for a bin $b$, if $\hat{\mathcal{J}}(b)>0$, a junction proposal is generated with its position computed by $\mathbf{x}_b + \hat{\mathcal{O}}(b)\cdot w$, where $\mathbf{x}_b$ is the position of the junction pixel, 
$\hat{\mathcal{O}}(b)$ is the learned offset vector, and $w$ is a rescaling factor of the offset.

\subsection{Line Segment and Junction Matching}\label{sec:matching}
\vspace{-2mm}
Line segment proposals and junction proposals are computed individually by leveraging different information, and their matching will provide more accurate meaningful alignment in wireframe parsing. We adopt a simple matching strategy to refining the initial proposals. A line segment proposal from the initial set is kept if and only if its two end-points can be matched with two junction proposals based on Euclidean distance with a predefined threshold $\tau$ ($\tau=10$ in all our experiments). A junction proposal will be removed if it does not match to any survived line segment proposal after refinement. After matching, line segments and junctions are coupled together, which will be further verified using a light-weight classifier.        

\subsection{Line Segment and Junction Verification}\label{sec:verification}
\vspace{-2mm}
Without loss of generality, let $\ddot{l}$ be a line segment proposal after refinement. A simple $2$-$fc$ layer is used as the validation head. To extract the same-sized feature vectors in $\mathcal{F}$ (the output of the feature backbone) for different line segments of different length for the head classifier, the widely used RoIPool/RoIAlign operation in the R-CNN based object detection system~\cite{fastRCNN, fasterrcnn} is adapted to line segments, and a simple LoIPool operation is used as done in L-CNN~\cite{ZhouQM19}. The LoIPool operation first uniformly samples $s$ points for a line segment $\ddot{l}$. The feature for each sampled point is computed from $\mathcal{F}$ using bi-linear interpolation as done in the RoIAlign operation and the 1D max-pooling operator is used to reduce the feature dimension. Then, all the features from the $s$ sampled points are concatenated as the feature vector for a line segment to be fed into the head classifier ($s=32$ in all our experiments).

In training the verification head classifier, we assign positive and negative labels to line segment proposals (after refinement) based on their distances to the groundtruth line segments. A line segment proposal is assigned to be a positive sample if there is a groundtruth line segment and their distance is less than a predefined threshold $\eta$ ($\eta=1.5$ in all our experiments). The distance between two line segments is computed as follows. We first match the two pairs of end-points based on the minimum Euclidean distance. Then, the distance between the two line segments is the maximum distance of the two endpoint-to-endpoint distances. So, the set of line segment proposals will be divided into the positive subset and the negative subset.

As illustrated in Fig.~\ref{teaser:learned-ls}, the negative subset usually contains many hard negative samples since the proposed holistic AFM usually generates line segment proposals of ``good quality", which is helpful to learn a better verification classifier. Apart from the learned positive and negative samples, we use a simple proposal augmentation method in a similar spirit to the static sampler used in L-CNN~\cite{ZhouQM19}: We add all the groundtruth line segments into the positive set. We also introduce a set of negative samples that are generated based on the groundtruth junction annotations (\ie,~line segments using the two end-points that do not correspond to any annotated line segment). 
During training, to avoid the class imbalance issue, we sample the same number, $n$, of positives and negatives (\ie, LoIs) from the two augmented subsets ($n=300$ in all our experiments).    
We use binary cross entropy loss in the verification module. Denote by $\mathbb{L}_{Ver}$ the loss computed on the sampled LoIs.

The proposed HAWP is trained end-to-end with the following loss function, 
\begin{equation}
    \mathbb{L} = \mathbb{L}_{LS} + \mathbb{L}_{Junc} + \mathbb{L}_{Ver}.
\end{equation}

\section{Experiments}\label{sec:exp}
\vspace{-2mm}
In this section, we present detailed experimental results and analyses of the proposed HAWP. \textit{Our reproducible PyTorch source code will be released}.

\textbf{Benchmarks.} The wireframe benchmark~\cite{Huang2018a} and the YorkUrban benchmark are used. The former consists of $5,000$ training samples and $462$ testing samples. The latter includes $102$ samples in total. The model is only trained on the former and tested on both. 

\textbf{Baselines.} Four methods are used: LSD~\cite{VonGioi2010}\footnote{The built-in LSD in OpenCV v3.2.0 is used in evaluation.}, AFM~\cite{afm}, DWP~\cite{Huang2018a}, and L-CNN~\cite{ZhouQM19} (the previous state-of-the-art approach).  The last three are DNN based approaches and the first one does not need training. The last two leverage junction information in training, and thus are directly comparable to the proposed HAWP. 

\textbf{Implementation Details.} To be fair in comparison with L-CNN, we adopt the same hyper-parameter settings (including those defined in Section~\ref{sec:HAWP}) when applicable in our HAWP. Input images are resized to $512\times 512$ in both training and testing. For the stacked Hourglass feature backbone, the number of stacks, the depth of each Hourglass module and the number of blocks are 2, 4, 1 respectively. Our HAWP is trained using the ADAM optimizer~\cite{adam} with a total of $30$ epochs on a single Tesla V100 GPU device. The learning rate, weight decay rate and batch size are set to $4\times 10^{-4}$, $ 1\times 10^{-4}$ and $6$ respectively. The learning rate is divided by $10$ at the $25$-th epoch. To further ensure apple-to-apple comparisons with L-CNN, we also re-train it using the same learning settings with slightly better performance obtained than those reported in their paper. 

\begin{table*}[t]
    \centering
    \resizebox{0.95\linewidth}{!}{ 
    \begin{tabular}{|l|ccc|c|c|l|l|ccc|c|c|l|l|l|}
        \hline
        \multirow{2}{*}{Method} & \multicolumn{7}{c|}{\textit{Wireframe Dataset}} & \multicolumn{7}{c|}{\textit{YorkUrban Dataset}}& \multirow{2}{*}{FPS}
        \\\cline{2-15}
        & sAP$^5$ & sAP$^{10}$ & sAP$^{15}$ & msAP & mAP$^{J}$ & AP$^{{H}}$ & F$^{{H}}$ & sAP$^5$ & sAP$^{10}$ & sAP$^{15}$ & msAP & mAP$^J$ & AP$^{{H}}$ & F$^{{H}}$  &
        \\\hline\hline
        LSD~\cite{VonGioi2010}& / & / & / & / & / & 55.2 & 62.5 &  / & / & / & / & / & 50.9 & 60.1 & \textbf{49.6}\\\hline 
        AFM~\cite{afm} & 18.5 & 24.4 & 27.5 & 23.5 & 23.3 & 69.2 & 77.2 & 7.3 & 9.4 & 11.1 & 9.3 & 12.4 &48.2 & 63.3 & 13.5 \\\hline
        DWP~\cite{Huang2018a} & 3.7 & 5.1 & 5.9 & 4.9 & 40.9 & 67.8 & 72.2  & 1.5 & 2.1 & 2.6 & 2.1 & 13.4 & 51.0 & 61.6 & 2.24\\\hline
        \multirow{2}{*}{L-CNN~\cite{ZhouQM19}} 
        & \multirow{2}{*}{58.9} & \multirow{2}{*}{62.9} & \multirow{2}{*}{64.9} & \multirow{2}{*}{62.2} & \multirow{2}{*}{59.3} & 80.3&  76.9 &  \multirow{2}{*}{24.3}& \multirow{2}{*}{26.4} & \multirow{2}{*}{27.5} & \multirow{2}{*}{26.1} & \multirow{2}{*}{30.4} &  58.5 & 61.8 & \multirow{2}{*}{15.6}\\
        &&&&&& 82.8$^\dag$ & 81.3$^\dag$
        &&&&&& 59.6$^\dag$ & 65.3$^\dag$    &
        \\\hline
                             
        \multirow{2}{*}{L-CNN (re-trained)} & \multirow{2}{*}{59.7} & \multirow{2}{*}{63.6} & \multirow{2}{*}{65.3} & \multirow{2}{*}{62.9} & \multirow{2}{*}{60.2} & 81.6 & 77.9 &  \multirow{2}{*}{25.0} & \multirow{2}{*}{27.1} & \multirow{2}{*}{28.3} & \multirow{2}{*}{26.8}&  \multirow{2}{*}{31.5}    &  58.3 & 62.2 & \multirow{2}{*}{15.6}\\
                               & &&&&& 83.7$^\dag$& 81.7$^\dag$ & & & & & & 59.3$^\dag$ & 65.2$^\dag$ & \\\hline
        \multirow{2}{*}{\textbf{HAWP (ours)}} & \multirow{2}{*}{\textbf{62.5}} & \multirow{2}{*}{\textbf{66.5}} & \multirow{2}{*}{\textbf{68.2}} &
        \multirow{2}{*}{\textbf{65.7}} & \multirow{2}{*}{\textbf{60.2}} & \textbf{84.5} & \textbf{80.3} &  \multirow{2}{*}{\textbf{26.1}} & \multirow{2}{*}{\textbf{28.5}} & \multirow{2}{*}{\textbf{29.7}} & \multirow{2}{*}{\textbf{28.1}} & 
        \multirow{2}{*}{\textbf{31.6}} &
        \textbf{60.6} & \textbf{64.8} & \multirow{2}{*}{\textit{29.5}}\\
                               & &&&&& 86.1$^\dag$& 83.1$^\dag$ &
                               &&&&& 61.2$^\dag$ & 66.3$^\dag$ & \\\hline
    \end{tabular}
    }
    
    \caption{Quantitative results and comparisons. Our propsed HAWP achieves state-of-the-art results consistently except for the FPS. The FPS of our HAWP is still significantly better than that of the three deep learning based methods. Note that for fair and apple-to-apple comparisons, we also retrained a L-CNN model using their latest released code and the same learning hyper-parameters used in our HAWP. Our retrained L-CNN obtained slightly better performance than the original one. $^{\dag}$ means that the post-processing scheme proposed in L-CNN~\cite{ZhouQM19} is used. {The FPS of L-CNN is computed without the post-processing.} See text for details.}
    \vspace{-3mm}
    \label{tab:summary-metric}
\end{table*}

\begin{figure*} [ht]
    \centering
    \includegraphics[width=0.245\linewidth]{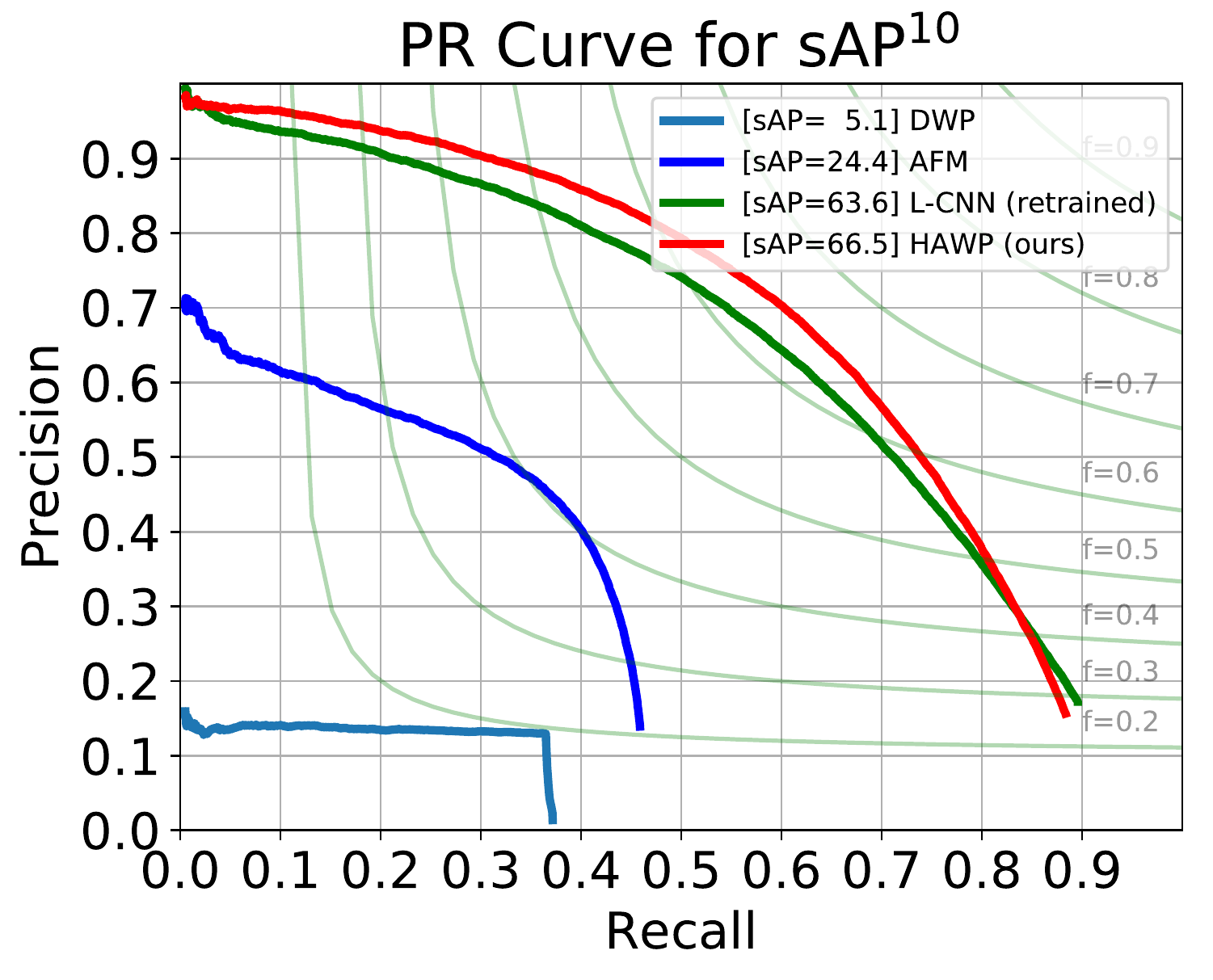} 
     \includegraphics[width=0.245\linewidth]{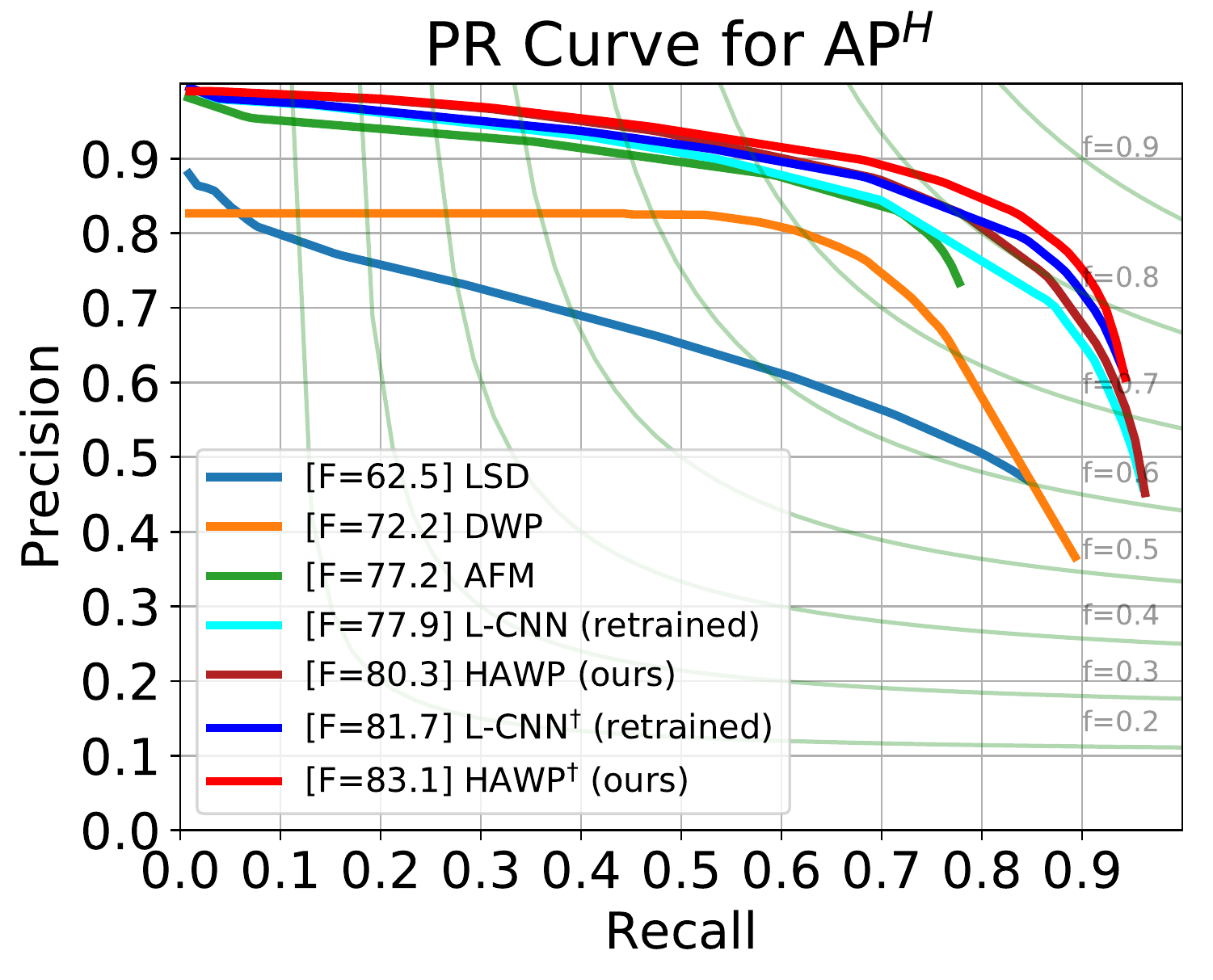}
      \includegraphics[width=0.245\linewidth]{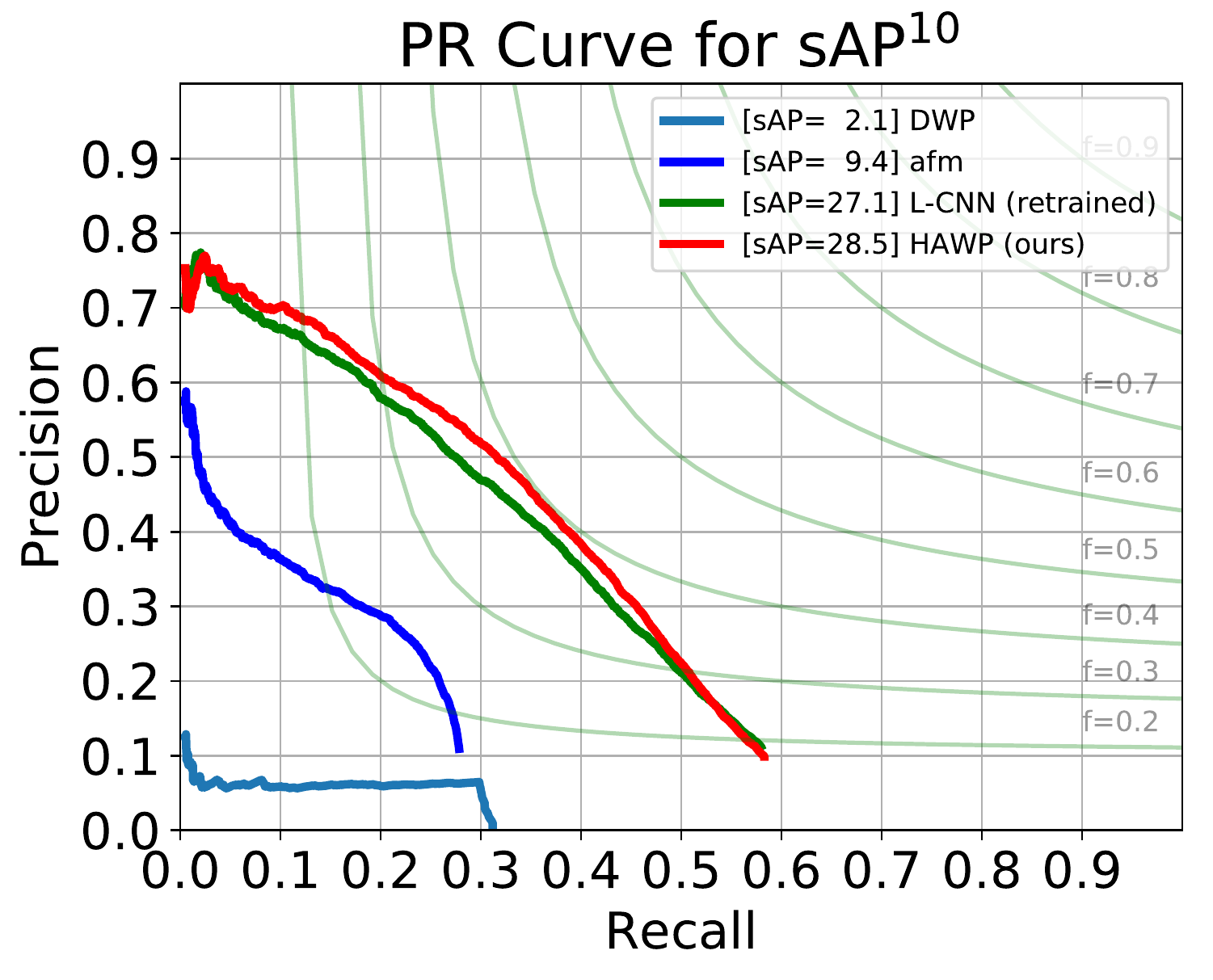}
      \includegraphics[width=0.245\linewidth]{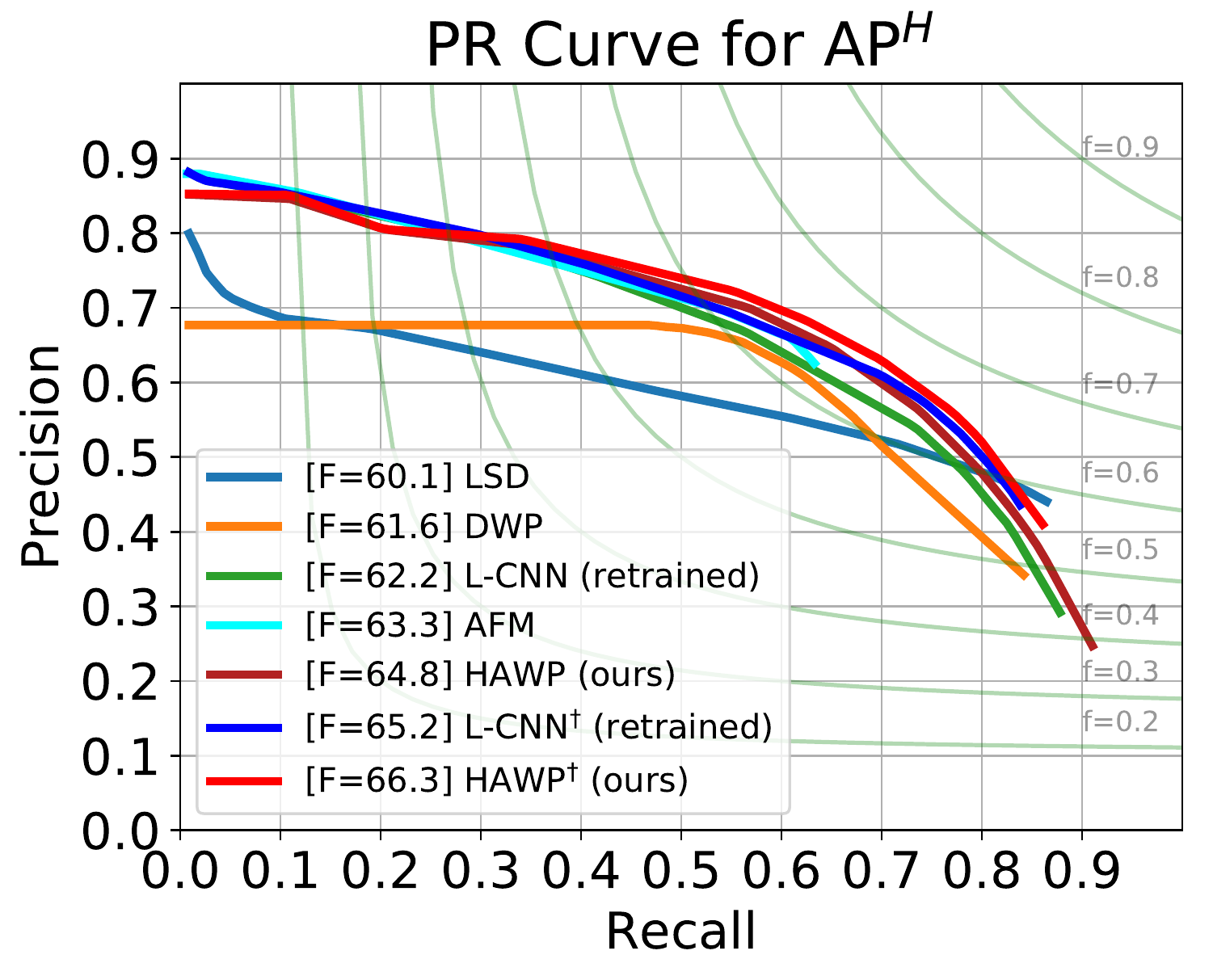}
    \caption{Precision-Recall (PR) curves of sAP$^{10}$ and AP$^{H}$ for DWP~\cite{Huang2018a}, AFM~\cite{afm}, L-CNN~\cite{ZhouQM19} and HAWP (ours) on the wireframe benchmark (the left two plots) and the YorkUrban benchmark (the right two plots). Best viewed in color and magnification.}
    \vspace{-3mm}
    \label{fig:pr-curves}
\end{figure*}

\subsection{Evaluation Metric}\label{sec:metric}
\vspace{-2mm} 
We follow the accuracy evaluation settings used in L-CNN  summarized as follows to be self-contained. 

\textbf{{Structural Average Precision (sAP)} of Line Segments}~\cite{ZhouQM19}. This is motivated by the typical AP metric used in evaluating object detection systems. A counterpart of the Intersection-over-Union (IoU) overlap is used. For each ground-truth line segment $\ddot{l}=(\mathbf{x}_1, \mathbf{x}_2)$, we first find the set of parsed line segments each of which, $\hat{\ddot{l}}=(\hat{\mathbf{x}}_1, \hat{\mathbf{x}}_2)$, satisfies the ``overlap", 
\begin{equation}
    \min_{(i,j)} \left\|\mathbf{x}_1 - \hat{\mathbf{x}}_i \right\|^2 + \left\|\mathbf{x}_2 - \hat{\mathbf{x}}_j \right\|^2 \leq \vartheta_{L},
\end{equation}
where $(i,j)=(1,2)$ or $(2,1)$, and $\vartheta_{L}$ is a predefined threshold. 
If the set of parsed line segments ``overlapping" with $\ddot{l}$ is empty, the line segment $\ddot{l}$ is counted as a False Negative (FN). If there are multiple candidates in the set, the one with the highest verification classification score is counted as a True Positive (TP), and the rest ones will be counted as False Positives (FPs).
A parsed line segment that does not belong to the candidate set of any groundtruth line segment is also counted as a FP. Then, sAP can be computed. 
To eliminate the influence of image resolution, the wireframe parsing results and the groundtruth wireframes are rescaled to the resolution of $128\times 128$ in evaluation. 
We set the threshold $\vartheta$ to $5, 10, 15$ and report the corresponding results, denoted by sAP$^{5}$, sAP$^{10}$, sAP$^{15}$. 
The overall performance of a wireframe parser is represented by the mean of the sAP values with different thresholds, denoted by \textit{msAP}.

\textbf{{Heatmap based F-score, F$^H$ and AP$^H$ of Line Segments.}} These are traditional metrics used in LSD and wireframe parsing~\cite{Huang2018a}. Instead of directly using the vectorized representation of line segments, heatmaps are used, which are generated by rasterizing line segments for both parsing results and the groundtruth. The pixel-level evaluation is used in calculating the precision and recall curves with which F$^H$ and AP$^H$ are computed.

\textbf{{Vectorized Junction Mean AP (mAP$^J$)}}~\cite{ZhouQM19}. It is computed in a similar spirit to msAP of line segments. Let $\vartheta_J$ be the thresold for the distance between a predicted junction and a groundtruth one. The mAP$^J$ is computed~\emph{w.r.t.}~$\vartheta_J=0.5, 1.0, 2.0$.

\begin{figure*}
    \def\outsize{0.138\linewidth}
    \centering
    \raisebox{35pt}{\rotatebox[origin=c]{90}{LSD}}
    \includegraphics[height=\outsize]{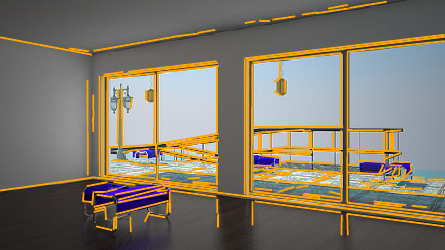}
    \includegraphics[height=\outsize]{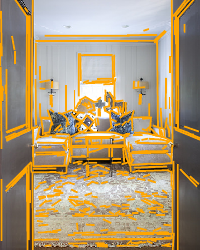}
    \includegraphics[height=\outsize]{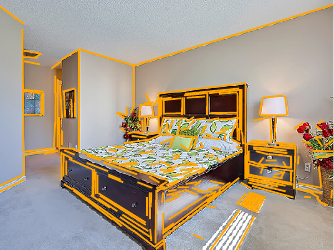}
    \includegraphics[height=\outsize]{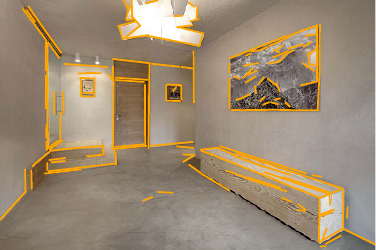}
    \includegraphics[height=\outsize]{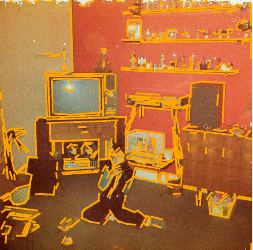} \\
    \raisebox{35pt}{\rotatebox[origin=c]{90}{DWP}}
    \includegraphics[height=\outsize]{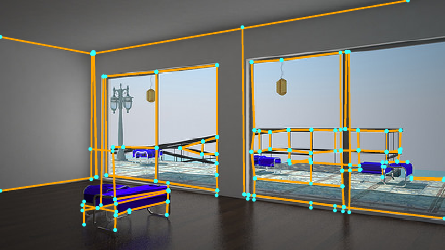}
    \includegraphics[height=\outsize]{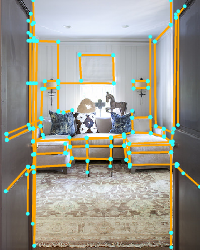}
    \includegraphics[height=\outsize]{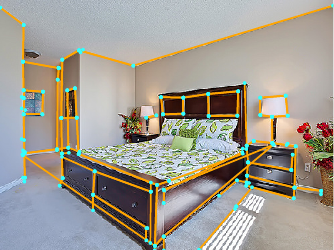}
    \includegraphics[height=\outsize]{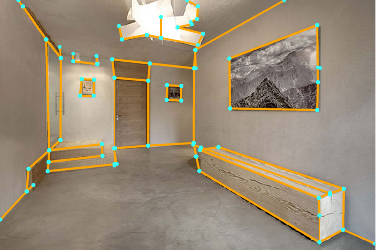}
    \includegraphics[height=\outsize]{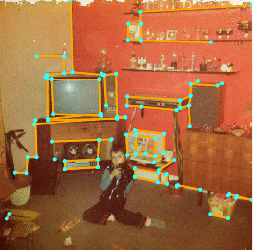} \\
    \raisebox{35pt}{\rotatebox[origin=c]{90}{AFM}}
    \includegraphics[height=\outsize]{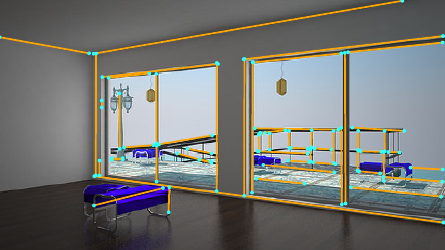}
    \includegraphics[height=\outsize]{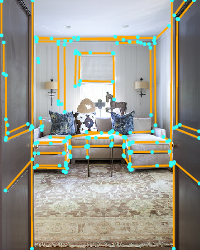}
    \includegraphics[height=\outsize]{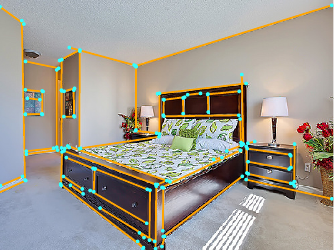}
    \includegraphics[height=\outsize]{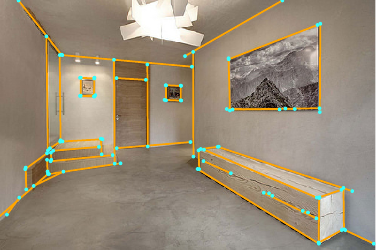}
    \includegraphics[height=\outsize]{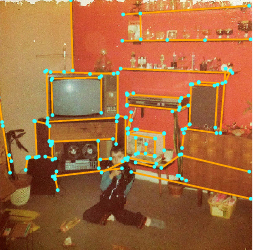} \\
    \raisebox{35pt}{\rotatebox[origin=c]{90}{L-CNN}}
    \includegraphics[height=\outsize]{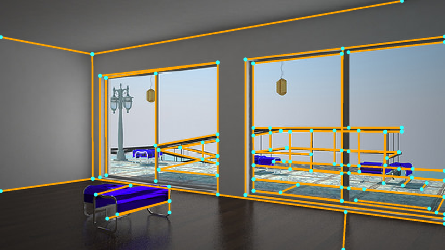}
    \includegraphics[height=\outsize]{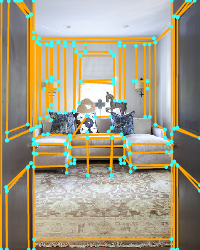}
    \includegraphics[height=\outsize]{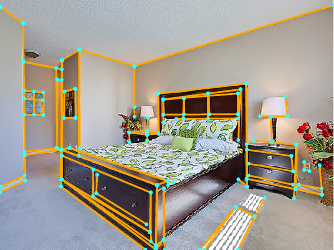}
    \includegraphics[height=\outsize]{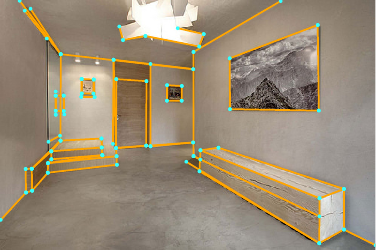}
    \includegraphics[height=\outsize]{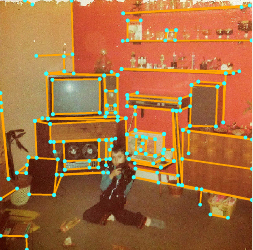} \\
    \hspace{-1.8pt}\raisebox{35pt}{\rotatebox[origin=c]{90}{HAWP (ours)}}
    \includegraphics[height=\outsize]{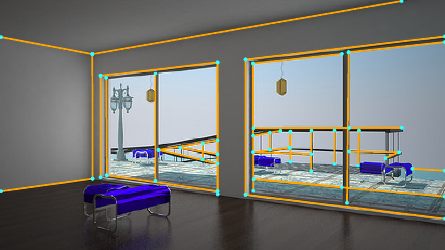}
    \includegraphics[height=\outsize]{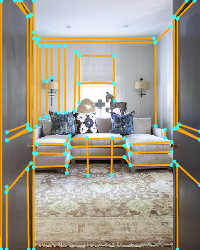}
    \includegraphics[height=\outsize]{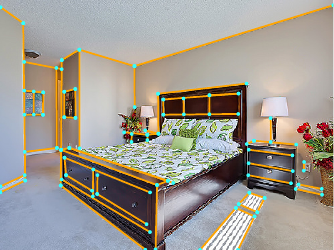}
    \includegraphics[height=\outsize]{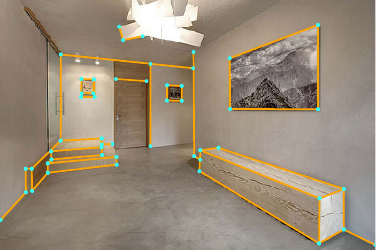}
    \includegraphics[height=\outsize]{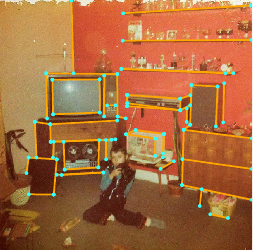} \\
    \raisebox{35pt}{\rotatebox[origin=c]{90}{\text{Ground Truth}}}
    \includegraphics[height=\outsize]{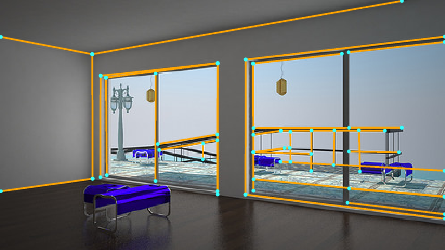}
    \includegraphics[height=\outsize]{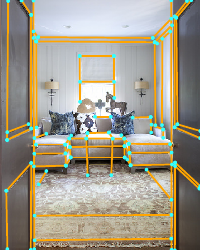}
    \includegraphics[height=\outsize]{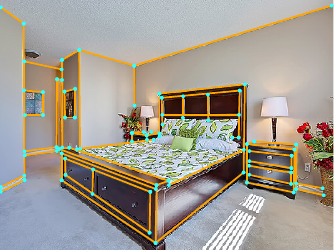}
    \includegraphics[height=\outsize]{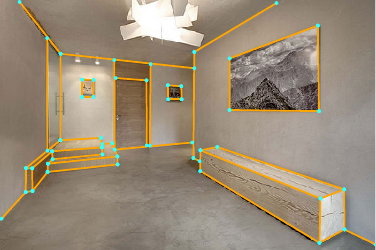}
    \includegraphics[height=\outsize]{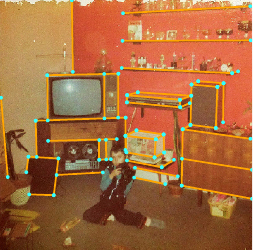}
    \caption{Wireframe parsing examples on the Wireframe dataset~\cite{Huang2018a}. 
    }
    \vspace{-4mm}
    \label{fig:visualization-wireframe}
\end{figure*}

\textbf{{Speed}}. Besides accuracy, speed is also important in practice. We use the frames per second (FPS) in evaluation. For fair comparisons, we compute the FPS for different methods under the same setting: the batch-size is $1$, and single CPU thread and single GPU (Tesla V100) are used. Note that the LSD~\cite{VonGioi2010} method does not take advantage of GPU.

\subsection{Results and Comparisons}\label{sec:results}
\vspace{-2mm}
\textbf{Quantitative Results.} Table~\ref{tab:summary-metric} summarizes the results  and comparisons in terms of the evaluation metric stated in Section~\ref{sec:metric}. \textbf{Our HAWP obtains state-of-the-art performance consistently}. In terms of the challenging msAP metric, it outperforms L-CNN by 2.8\% and 1.3\% (absolute improvement) on the wireframe benchmark and the YorkUrban benchmark respectively. It also runs much faster than  L-CNN  with 89\% relative improvement in FPS. AFM and DWP are relatively slow due to their non-GPU friendly post-processing modules entailed for performance. In terms of the heatmap based evaluation metric, our  HAWP is also significantly better than L-CNN regardless of the post-processing module proposed in L-CNN. Fig.~\ref{fig:pr-curves} shows comparisons of PR curves.  

Since our proposed HAWP and L-CNN use very similar wireframe parsing pipelines and adopt the same design choices when applicable. \textit{The consistent accuracy gain of our HAWP must be contributed by the novel 4-D holistic attraction field representation and its integration in the parsing pipeline.} In terms of efficiency, our HAWP runs much faster since a significantly fewer number of line segment proposals are used in the verification module. As shown in Table~\ref{tab:profile}, our HAWP uses 5.5 times fewer number of line segment proposals.

\begin{table}[]
    \centering
    \resizebox{0.95\linewidth}{!}{
    \begin{tabular}{|c|c|c|c|c|c|}
    \hline
                 & \# Junctions & \# Proposals & sAP$^{10}$ & FPS & \# GT Lines\\\hline \hline
         L-CNN~\cite{ZhouQM19}   & 159.2 & 22k & 63.6 & 15.6 & \multirow{2}{*}{74.2}\\
         HAWP (ours)    & 189.6 & \textbf{4k} & \textbf{66.5} & \textbf{29.5} &\\\hline
    \end{tabular}
    }
    \caption{Performance profiling on the Wireframe dataset. \#Proposals represents the number of line segments in verification. The average number of groundtruth is listed in the last row.
    }
    \label{tab:profile}
\end{table}

\textbf{Qualitative Results.} Fig.~\ref{fig:visualization-wireframe} shows wireframe parsing results by the five methods.

\subsection{Ablation Study}\label{sec:ablation-sampling}
\vspace{-2mm}
We compare the effects of three aspects: our proposed H-AFM \textit{vs.} the vanilla AFM~\cite{afm}, the distance residual module (Section~\ref{sec:lsd}), and the composition of negative samples in training verification module (Section~\ref{sec:verification}).

\begin{table}[ht]
    \centering
    \vspace{-2mm}
    \resizebox{0.95\linewidth}{!}{
    \begin{tabular}{|l|cc|cccc|ccc|}
        \hline
         \multirow{2}{*}{\makecell{
         Line Segment\\Representation}}  &   \multicolumn{2}{c|}{Distance Residual}  & \multicolumn{4}{c|}{Negative Example Sampler}  & \multicolumn{3}{c|}{Performance}\\
                                                        &
                                                        Training& Testing
                                                        &
         $\mathbb{N}^*$ & $\mathbb{D}^*$ & $\mathbb{D}^-$ & $\mathbb{S}^-$
         & sAP$^5$ & sAP$^{10}$ & sAP$^{15}$\\\hline
         \multirow{4}{*}{H-AFM} & \multirow{4}{*}{\pmb{\checkmark}} & \multirow{4}{*}{\pmb{\checkmark}} & \checkmark &   &   & \checkmark & \textbf{62.5} & \textbf{66.5} & \textbf{68.2}\\ \cline{4-10}
                               &                      &                      & \checkmark &   &   &    & 62.0 & 66.0 & 67.6\\ \cline{4-10}
                               &                      &                      &    & \checkmark& \checkmark& \checkmark & 62.2 & 66.1 & 67.8\\ \cline{4-10} 
                               &                      &                      &    &   & \checkmark& \checkmark & 62.0 & 65.8 & 67.4\\\hline
         H-AFM & \checkmark &    & \checkmark &   &   & \checkmark & 58.9 & 63.0 & 64.8 \\ \cline{2-10}
         H-AFM &   &    & \checkmark &   &   & \checkmark & 58.7 & 62.6 & 64.4 \\\hline
         ~~~~AFM  &   &    & \checkmark &   &   & \checkmark & 30.9 & 33.7 & 35.0\\\hline
    \end{tabular}
    }
    
    \caption{The ablation study of three design and learning aspects in the proposed HAWP. See text for details.}
    \vspace{-1mm}
    \label{tab:ablation-study}
\end{table}

Table~\ref{tab:ablation-study} summarizes the comparisons. We observe that both H-AFM and the distance residual module are important for improving performance. 
The natural negative sampler $\mathbb{N}^*$ randomly chooses negative line segments based on the matching results (with respect to the annotations).
The rest of three negative example samplers ($\mathbb{D}^*, \mathbb{D}^-, \mathbb{S}^-$) are also investigated in L-CNN and their full combination is needed for training L-CNN.  $\mathbb{D}^*$ randomly selects a part of examples from the online generated line segment proposals, regardless of the matching results. $\mathbb{D}^-$ tries to match the proposals with pre-computed hard negative examples and the matched proposals are used as negative samples. $\mathbb{S}^-$ directly obtains the negative examples from the pre-computed hard negative examples set. 
In our experiment, the number of samples for $\mathbb{N}^*, \mathbb{D}^*, \mathbb{D}^-$ and $\mathbb{S}^-$ are set to $300, 300, 300, 40$ respectively. 
We observe that our HAWP is less sensitive to those samplers due to the informative line segment proposal generation stage.  

\section{Conclusions and Discussions}\vspace{-2mm}
This paper presents a holistically-attracted wireframe parser (HAWP) with state-of-the-art performance obtained on two benchmarks, the wireframe dataset and the YorkUrban dataset. The proposed HAWP consists of three components: proposal (line segments and junctions) initialization, proposal refinement and proposal verification, which are end-to-end trainable. Compared to the previous state-of-the-art wireframe parser L-CNN~\cite{ZhouQM19}, our HAWP is enabled by a novel 4-D holistic attraction field map representation  (H-AFM) for line segments in proposal generation stages. Our HAWP also achieves real-time speed with a single GPU, and thus is useful for many downstream tasks such as SLAM and Structure from Motion (SfM). The proposed H-AFM is also potentially useful for generic LSD problems in other domains such as medical image analysis.

{\small
\bibliographystyle{ieee_fullname}
\bibliography{egbib}
}

\end{document}